\pdfoutput=1

\documentclass[11pt]{article}

\usepackage{acl}

\usepackage{times}
\usepackage{latexsym}
\usepackage[splitrule]{footmisc}

\usepackage[T1]{fontenc}

\usepackage[utf8]{inputenc}

\usepackage{microtype}

\usepackage{graphicx} 

\title{Task Transfer and Domain Adaptation for Zero-Shot Question Answering}

\author{Xiang Pan$^{*}$ \\
  New York University \\
  \texttt{xiangpan@nyu.edu} \\\And
  Alex Sheng$^{*}$ \\
  New York University \\
  \texttt{alexsheng4@gmail.com} \\\And
  David Shimshoni$^{*}$ \\
  New York University \\
  \texttt{ds5396@nyu.edu} \\\AND
  Aditya Singhal$^{*}$ \\
  New York University \\
  \texttt{adis@nyu.edu} \\\And
  Sara Rosenthal \\
  IBM Research AI \\
  \texttt{sjrosenthal@us.ibm.com} \\\And
  Avirup Sil \\
  IBM Research AI \\
  \texttt{avi@us.ibm.com}
  }

\begin{document}
\maketitle

\def\thefootnote{*}\footnotetext{Equal Contribution}\def\thefootnote{\arabic{footnote}}

\begin{abstract}
Pretrained language models have shown success in various areas of natural language processing, including reading comprehension tasks. However, when applying machine learning methods to new domains, labeled data may not always be available. To address this, we use supervised pretraining on source-domain data to reduce sample complexity on domain-specific downstream tasks. We evaluate zero-shot performance on domain-specific reading comprehension tasks by combining task transfer with domain adaptation to fine-tune a pretrained model with no labelled data from the target task. Our approach outperforms Domain-Adaptive Pretraining on downstream domain-specific reading comprehension tasks in 3 out of 4 domains.
\end{abstract}

\begin{figure*}[t]
\centering
\includegraphics[scale=0.3]{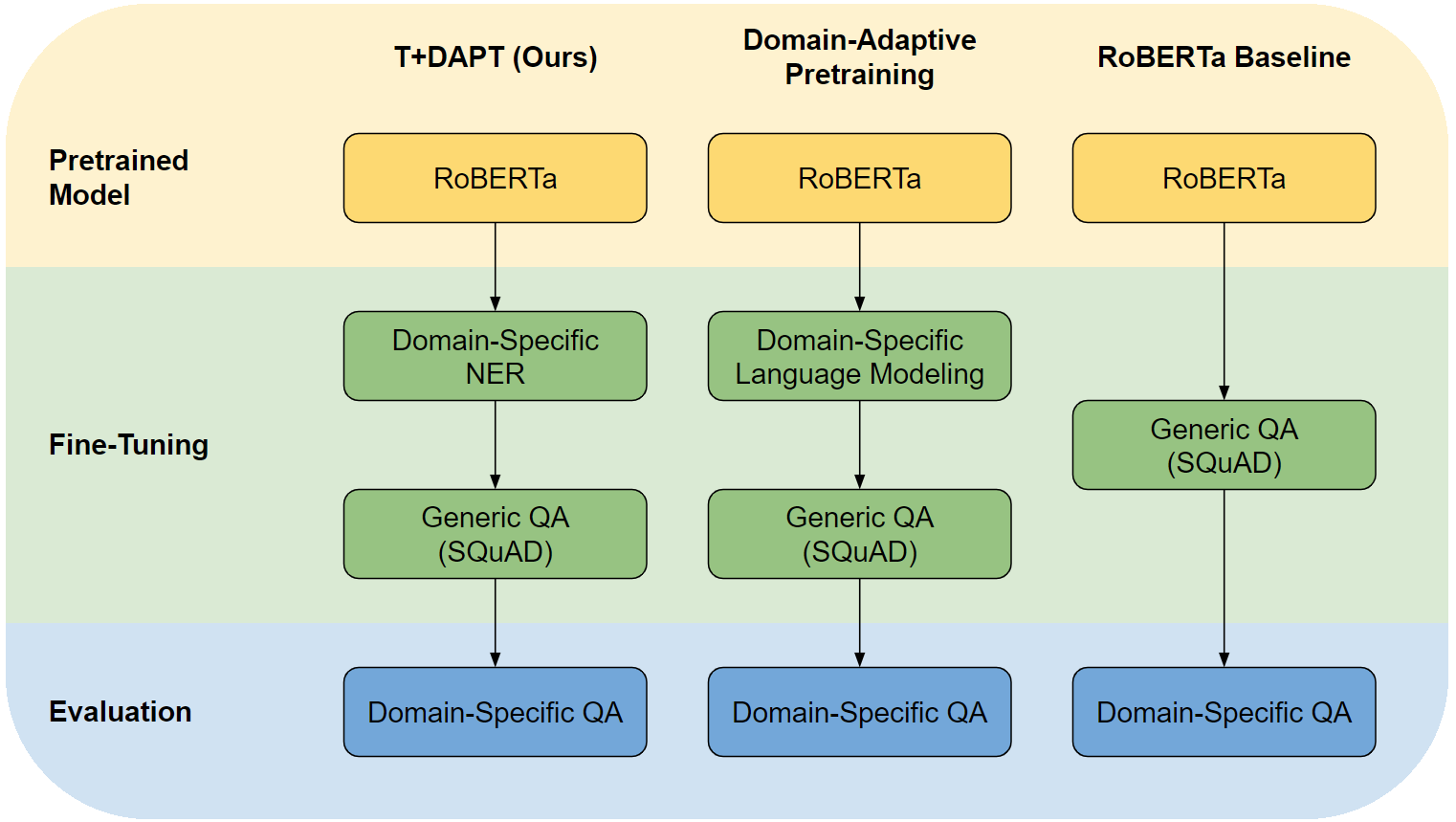}
\caption{sequential transfer learning procedures of T+DAPT, DAPT, and a RoBERTa baseline for zero-shot question answering.} \label{fig:framework}
\end{figure*}

\section{Introduction}

Pretrained language models \cite{liu_2019_roberta, wolf2020huggingfaces} require substantial quantities of labeled data to learn downstream tasks. For domains that are novel or where labeled data is in short supply, supervised learning methods may not be suitable \cite{zhang_2020_multistage, madasu_2020_sequential, rietzler-etal-2020-adapt}. Collecting sufficient quantities of labeled data for each new application can be resource intensive, especially when aiming for both a specific task type and a specific data domain. By traditional transfer learning methods, it is prohibitively difficult to fine-tune a pretrained model on a domain-specific downstream task for which there is no existing training data. In light of this, we would like to use more readily available labeled in-domain data from unrelated tasks to domain-adapt our fine-tuned model.

In this paper, we consider a problem setting where we have a domain-specific target task (QA) for which we do not have any in-domain training data (QA Data in the target domain). However, we assume that we have generic training data for the target task type, and in-domain training data for another task. To address this problem setting, we present Task and Domain Adaptive Pretraining (T+DAPT), a technique that combines domain adaptation and task adaptation to improve performance in downstream target tasks. We evaluate the effectiveness of T+DAPT in zero-shot domain-specific machine reading comprehension (MRC) \cite{hazen_2019_towards, reddy2020endtoend,  wiese_2017_neural} by pretraining on in-domain NER data and fine-tuning for generic domain-agnostic MRC on SQuADv1 \cite{rajpurkar_2016_squad}, combining knowledge from the two different tasks to achieve zero-shot learning on the target task. We test the language model’s performance on domain-specific reading comprehension data taken from 4 domains: News, Movies, Biomedical, and COVID-19. In our experiments, RoBERTa-Base models trained using our approach perform favorably on domain-specific reading comprehension tasks compared to baseline RoBERTa-Base models trained on SQuAD as well as Domain Adaptive Pretraining (DAPT). Our code is publicly available for reference. \footnote[1]{\url{https://github.com/adityaarunsinghal/Domain-Adaptation}}

We summarize our contributions as follows:
\begin{itemize}
    \setlength{\itemsep}{0pt}
    \setlength{\parsep}{0pt}
    \setlength{\parskip}{0pt}
    \item We propose Task and Domain Adaptive Pretraining (T+DAPT) combining domain adaptation and task adaptation to achieve zero-shot learning on domain-specific downstream tasks.
    \item We experimentally validate the performance of T+DAPT, showing our approach performs favorably compared to both a previous approach (DAPT) and a baseline RoBERTa fine-tuning approach.
    \item We analyze the adaptation performance on different domains, as well as the behavior of DAPT and T+DAPT under various experimental conditions.
\end{itemize}

\section{Related Work}

It has been shown that pretrained language models can be domain-adapted with further pretraining \cite{pruksachatkun_intermediate} on unlabeled in-domain data to significantly improve the language model's performance on downstream supervised tasks in-domain. This was originally demonstrated by BioBERT \citep{lee_2019_biobert}. \citet{gururangan_2020_dont} further explores this method of domain adaptation via unsupervised pretraining, referred to as Domain-Adaptive Pretraining (DAPT), and demonstrates its effectiveness across several domains and data availability settings. This procedure has been shown to improve performance on specific domain reading comprehension tasks, in particular in the biomedical domain \citep{gu2021domainspecific}. In this paper, as a baseline for comparison, we evaluate the performance of DAPT-enhanced language models in their respective domains, both in isolation with SQuAD1.1 fine-tuning and in conjunction with our approach that incorporates the respective domain’s NER task. DAPT models for two of our domains, News and Biomedical, are initialized from pretrained weights as provided by the authors of \citet{gururangan_2020_dont}. We train our own DAPT baselines on the Movies and COVID-19 domains.  \citet{xu_2020_forget} explore methods to reduce catastrophic forgetting during language model fine-tuning. They apply topic modeling on the MS MARCO dataset \citep{bajaj_2018_ms} to generate 6 narrow domain-specific data sets, from which we use BioQA and MoviesQA as domain-specific reading comprehension benchmarks.

\section{Experiments}

\begin{table*}[phbt]
\setlength{\tabcolsep}{3pt}
\centering
\begin{tabular}{l l p{0.8\linewidth}}
\hline
\textbf{Dataset} & \textbf{Dev Set} & \textbf{Sample} \\
\hline
MoviesQA & 755 & Q: After its re-opening, which types of movies did the Tower Theatre show? \\
 & & A: second and third run movies, along with classic films \\
\hline
NewsQA & 934 & Q: Who is the struggle between in Rwanda? \\
 & & A: The struggle pits ethnic Tutsis, supported by Rwanda, against ethnic Hutu, backed by Congo. \\
\hline
BioQA & 4,790 & Q: What is hemophilia? \\
 & & A: a bleeding disorder characterized by low levels of clotting factor proteins. \\
\hline
CovidQA & 2,019 & Q: What is the molecular structure of bovine coronavirus? \\
 & & A: single-stranded, linear, and nonsegmented RNA \\
\hline
\end{tabular}
\caption{Overview of the domain-specific MRC datasets used in our experiments. The number of question-answer pairs in the train set and development set for each domain is shown, along with a sample question-answer pair from each domain. The datasets share the same format as SQuAD.}
\label{table:dataset-overview}
\end{table*}

We aim to achieve zero-shot learning for an unseen domain-specific MRC task by fine-tuning on both a domain transfer task and a generic MRC task. The model is initialized by pretrained RoBERTa weights \cite{liu_2019_roberta}, then fine-tuned using our approach with a domain-specific supervised task to augment domain knowledge, and finally trained on SQuAD to learn generic MRC capabilities to achieve zero-shot MRC in the target domain on an unseen domain-specific MRC task without explicitly training on the final task. This method is illustrated in Figure \ref{fig:framework}.

\subsection{Datasets}
We explore the performance of this approach in the Movies, News, Biomedical, and COVID-19 domains. Specifically, our target domain-specific MRC tasks are MoviesQA  \citep{xu_2020_forget}, NewsQA \citep{trischler_2017_newsqa}, BioQA \citep{xu_2020_forget}, and CovidQA \citep{mller_2020_covidqa}, respectively. We choose to use named entity recognition (NER) as our supervised domain adaptation task for all four target domains, as labeled NER data is widely available across various domains. Furthermore, NER and MRC share functional similarities, as both rely on identifying key tokens in a text as entities or answers. The domain-specific NER tasks are performed using supervised training data from the MIT Movie Corpus \citep{liu_2013_query}, CoNLL 2003 News NER  \citep{tjongkimsang_2003_introduction}, NCBI-Disease \citep{doan_2014_ncbi} and COVID-NER \footnote[2]{\url{https://github.com/tsantosh7/COVID-19-Named-Entity-Recognition}}. The domain-specific language modeling tasks for DAPT are performed using unsupervised text from IMDB \citep{maas_2011_learning}, the RealNews Corpus \citep{zellers_defending}, the Semantic Scholar Open Research Corpus \citep{lo2020s2orc} and the Covid-19 Corpus \footnote[3]{\url{https://github.com/davidcampos/covid19-corpus}}.

\subsection{Methods}
We compare our approach (T+DAPT) to a previous approach (DAPT) as well as a baseline model. For the baseline, the pretrained RoBERTa-Base model is fine-tuned on SQuAD and evaluated on domain-specific MRC without any domain adaptation. In the DAPT approach, RoBERTa-Base is first initialized with fine-tuned DAPT weights (NewsRoBERTa and BioRoBERTa) provided by \citet{gururangan_2020_dont} or implemented ourselves using the methodology described in \citet{gururangan_2020_dont} and different Movies and COVID-19 datasets \citep{maas_2011_learning, danescuniculescumizil_2011_chameleons, pang_2019_thumbs}. These models are initialized by DAPT weights---which have been fine-tuned beforehand on unsupervised text corpora for domain adaptation---from the HuggingFace model hub \cite{wolf2020huggingfaces}, fine-tuned on SQuAD, and evaluated on domain-specific MRC.
\subsection{Results}
We compare the effectiveness of our approach, which uses NER instead of language modeling (as in DAPT) for the domain adaptation method in a sequential training regime. Our experiments cover every combination of domain (Movies, News, Biomedical, or COVID) and domain adaptation method (T+DAPT which uses named entity recognition vs. DAPT which uses language modeling vs. baseline with no domain adaptation at all).

\begin{table*}[tp]
\centering
\begin{tabular}{lllll}
\hline
\textbf{RoBERTa Retraining Procedure} & \textbf{MoviesQA} & \textbf{NewsQA} & \textbf{BioQA} & \textbf{CovidQA} \\
\hline
SQuAD1.1 & 67.1 & \textbf{57.0} & 58.0 & 42.0 \\
DAPT + SQuAD1.1 & 60.7 & 54.4 & 57.8 & \textbf{47.2} \\
\textit{T+DAPT} (ours) & \textbf{68.0} & 56.0 & \textbf{58.9} & 42.7 \\
DAPT + \textit{T+DAPT} & 66.4 & 54.2 & 55.1 & 43.1 \\
\hline
\end{tabular}
\caption{
F1 score of pretrained RoBERTa-Base models on dev sets of MRC datasets for given domains with the stated retraining regimens}
\label{tab:main-result}
\end{table*}

Our results are presented in Table \ref{tab:main-result}. We use F1 score to evaluate the QA performance of each model in its target domain. In our experiments, DAPT performs competitively with baseline models and outperforms in one domain (CovidQA). Our T+DAPT approach (RoBERTA + Domain NER + SQuAD) outperforms the baseline in three out of four domains (Movies, Biomedical, COVID) and outperforms DAPT in three out of four domains (Movies, News, Biomedical). We also test a combination of DAPT and T+DAPT by retraining DAPT models on domain NER then SQuAD, and find that this combined approach underperforms compared to either T+DAPT alone or DAPT alone in all four domains. We further discuss the possible reasons for these results in Section \ref{analysis}.

\section{Analysis}
\label{analysis}

\textbf{Specific domains learn from adaptation:} Our approach shows promising performance gains when used for zero-shot domain-specific question answering, particularly in the biomedical, movies, and COVID domains, where the MRC datasets were designed with the evaluation of domain-specific features in mind. Performance gains are less apparent in the News domain, where the NewsQA dataset was designed primarily to evaluate causal reasoning and inference abilities---which correlate strongly with SQuAD and baseline RoBERTa pretraining---rather than domain-specific features and adaptation. The lack of performance gains from either T+DAPT or DAPT in the News domain could also possibly be attributed to the nature of the domain: \citet{gururangan_2020_dont} found that the News domain had the highest vocabulary overlap of any domain (54.1\%) with the RoBERTa pretraining corpus, so the baseline for this domain could have had an advantage in the News domain that would be lost due to catastrophic forgetting while little relevant knowledge is gained from domain adaptation. We perform follow-up experiments with varying amounts of epochs and training data in SQuAD fine-tuning to analyze the tradeoff between more thorough MRC fine-tuning and better preservation of source domain knowledge from DAPT and auxiliary domain adaptation tasks. The results from these runs are in the Appendix (Table \ref{table:additional-result}).

\textbf{When does DAPT succeed or fail:} In zero-shot QA, DAPT performs competitively with the baseline in all domains and outperforms in the COVID domain. This builds upon the results of \citet{gururangan_2020_dont}, which reports superior performance on tasks like relation classification, sentiment analysis, and topic modeling, but does not address reading comprehension tasks, which DAPT may not have originally been optimized for. Unsupervised language modeling may not provide readily transferable features for reading comprehension, as opposed to NER which identifies key tokens and classifies those tokens into specific entities. These entities are also often answer tokens in reading comprehension, lending to transferable representations between NER and reading comprehension. Another possible factor is that RoBERTa was pretrained on the English Wikipedia corpus, the same source that the SQuAD questions were drawn from. Because of this, it is possible that pretrained RoBERTa already has relevant representations that would provide an intrinsic advantage for SQuAD-style reading comprehension which would be lost due to catastrophic forgetting after retraining on another large language modeling corpus in DAPT.

In the COVID domain, we use the article dataset from \citet{wang2020cord}. These articles also make the basis for the CovidNER and CovidQA \cite{moller-etal-2020-covid} datasets, which may explain the large performance improvement from DAPT in this domain. These results suggest that the performance of DAPT is sensitive to the similarity of its language modeling corpus to the target task dataset.\footnote{Additional experiments in the COVID domain with different auxiliary tasks are presented in the Appendix \ref{app:addi}}

\begin{table*}[phbt]
\centering
\begin{tabular}{p{0.95\linewidth}}
\hline
\textbf{BioQA Samples} \\
\hline
Q: what sugar is found in rna \\
DAPT: ribose, whereas the sugar in DNA is deoxyribose \\
T+DAPT: ribose \\
\hline
Q: normal blood pressure range definition \\
DAPT: 120 mm Hg1 \\
T+DAPT: a blood pressure of 120 mm Hg1 when the heart beats (systolic) and a blood pressure of 80 mm Hg when the heart relaxes (diastolic) \\
\hline
\textbf{MoviesQA Samples} \\
\hline
Q: what is cyborgs real name \\
DAPT: Victor Stone/Cyborg is a hero from DC comics most famous for being a member of the Teen Titans \\
T+DAPT: Victor Stone \\
\hline
Q: who plays klaus baudelaire in the show \\
DAPT: Liam Aiken played the role of Klaus Baudelaire in the 2004 movie A Series of Unfortunate Events. \\
T+DAPT: Liam Aiken \\
\hline
\end{tabular}
\caption{Samples from BioQA and MoviesQA where T+DAPT achieves exact match with the label answer, and DAPT produces a different answer. Answers from each approach are shown side-by-side for comparison.}
\end{table*}
\label{table:comparison-samples}

\section{Conclusion}

We evaluate the performance of our T+DAPT approach with domain-specific NER, achieving positive results in a zero-shot reading comprehension setting in four different domain-specific QA datasets. These results indicate that our T+DAPT approach robustly improves performance of pretraining language models in zero-shot domain QA across several domains, showing that T+DAPT is a promising approach to domain adaptation in low-resource settings for pretrained language models, particularly when directly training on target task data is difficult.

In future work, we intend to explore various methods to improve the performance of T+DAPT by remedying catastrophic forgetting and maximizing knowledge transfer. For this we hope to emulate the regularization used by \citet{xu_2020_forget} and implement multi-task learning and continual learning methods like AdapterNet \citep{hazan2018adapternet}. In order to improve the transferability of learned features, we will explore different auxiliary tasks such as NLI and sentiment analysis in addition to few-shot learning approaches.

\section{Ethical Considerations}

Question answering systems are useful tools in complement to human experts, but the “word-of-machine effect” \citep{word-machine} demonstrates the effects of a potentially dangerous over-trust in the results of such systems. While the methods proposed in this paper would allow more thorough usage of existing resources, they also bestow confidence and capabilities to models which may not have much domain expertise. T+DAPT models aim to mimic extensively domain-trained models, which are themselves approximations of real experts or source documents. Use of domain adaptation methods for low-data settings could propagate misinformation from a lack of source data. For example, while making an information-retrieval system for biomedical and COVID information could become quicker and less resource-intensive using our approach, people should not rely on such a system for medical advice without extensive counsel from a qualified medical professional.

\section*{Acknowledgement}

We thank Sam Bowman for providing key feedback throughout our research.

\bibliography{anthology,custom}
\bibliographystyle{acl_natbib}

\clearpage

\appendix

\section{Appendix}

\label{sec:appendix}

\begin{table}[phbt]
\centering
\begin{tabular}{lllll}
\hline
\textbf{RoBERTa Adaptation Procedure} &  \textbf{CovidQA} \\
\hline
CovidQA (upper bound) & 52.1416 \\
SQuAD only & 42.0485 \\
DAPT  & 47.2190 \\
CovidNER  & 42.6584 \\
CovidQCLS  & 42.6300 \\
DAPT+Covid-NER  & 43.0710 \\
DAPT+Covid-QCLS  & \textbf{45.8314} \\
DAPT+CovidNER+CovidQCLS  & 43.0854 \\
\hline
\end{tabular}

\caption{
Zero-shot F1 performance of RoBERTa-Base models on dev sets of QA data for COVID domain with SQuAD1.1 following different intermediate pretraining regimens. The CovidQA upper bound score is attained by training directly on the CovidQA train set.}
\label{table:additional-result}
\end{table}

\begin{table}[phbt] 
\centering
\begin{tabular}{lllll}
\hline
\textbf{Model} & \textbf{NewsQA} \\
\hline
RoBERTa-Base \\
\hline
1 Epoch, 1000 Samples & 19.9953 \\
2 Epochs, 1000 Samples & 35.2666 \\
2 Epochs, 5000 Samples & 47.0090 \\
2 Epochs, All Samples & \textbf{56.9803} \\
2 Epochs, All Samples (Head) & 05.5891 \\
\hline
NewsRoBERTa (DAPT) \\
\hline
1 Epoch, 1000 Samples  & 17.9025 \\
2 Epochs, 1000 Samples  & 28.4453 \\
2 Epochs, 5000 Samples & 44.1206 \\
\hline

\end{tabular}

\caption{
Zero-shot F1 performance of RoBERTa-Base models on NewsQA following different amounts of SQuAD fine-tuning. For comparison the score of our News model from the main paper with 2 epochs and all samples is included as an upper bound, alongside a head tuning baseline where all weights are frozen except the classifier layer.}
\label{table:comparison-samples}
\end{table}


\subsection{Experiment Details and Additional Experiments}\label{app:addi}

\textbf{Freezing Layer} - 
We tried to freeze the bottom layer after NER training and only train the QA layer on SQuAD, the performance is worse than fine-tuning the whole RoBERTa and QA layer. NER and QA may not rely on the exact same features for the final task which may be the reason that freezing causes a performance decrease.

\textbf{Different Training Epoch and Training Examples} - 
When selecting the best performance model, we use a validation set in target domain to evaluate the performance. From Table \ref{table:comparison-samples}, we show our trials with different amounts of SQuAD training in the News Domain and how it affected performance in NewsQA. 

\textbf{Different Training Order} - 
We tried to use different training order, for example, we train on SQuAD1.1 task first and then on NER, the F1 score is 42.15 in CovidQA, which has some improvement, but QA as the last task performs better.

\textbf{Another Auxiliary Task} - 
In the Covid domain, we also do experiments on a more QA-relevant task, question classification (QCLS) \cite{wei2020people}. We show the result in Table \ref{table:additional-result}. The experiments show that QCLS task have more improvements than NER task. In addition, we test the model trained on CovidQA as the performance upper bound. 

\end{document}